\documentclass[10pt,twocolumn,letterpaper]{article}

\usepackage[pagenumbers]{cvpr} % To force page numbers, e.g. for an arXiv version

\definecolor{cvprblue}{rgb}{0.21,0.49,0.74}
\usepackage[pagebackref,breaklinks,colorlinks,allcolors=cvprblue]{hyperref}

\def\confName{CVPR}
\def\confYear{2026}

\title{A Robust Semantic Segmentation Pipeline for the CVPR 2026\\ 8th UG2+ Challenge Track 2}

\author{
Jinming Chai, Libo Yan, Licheng Jiao, Fang Liu\\
School of Artificial Intelligence, Xidian University, Xi’an 710071, China\\
{\tt\small jinmingchai@163.com, liboyan28@163.com, lchjiao@mail.xidian.edu.cn, f63liu@163.com}
}

\begin{document}
\maketitle
\begin{abstract}
This report presents our solution for the WeatherProof Dataset Challenge, namely CVPR 2026 8th UG2+ Challenge Track 2: Semantic Segmentation in Adverse Weather. For the semantic segmentation task under adverse weather conditions, we propose a semi-supervised segmentation pipeline. Our method is trained exclusively on the WeatherProof dataset, without using any additional external data. Specifically, we adopt UniMatch V2 as the baseline model and treat all degraded-weather images as unlabeled data for semi-supervised training, thereby fully exploiting the data distribution provided by the challenge. During inference, we further apply test-time augmentation to improve the robustness and segmentation accuracy of the final predictions. The code is publicly available at: \url{https://github.com/ylb888/weatherproof-challenge-unimatchv2}.
\end{abstract}    
\section{Introduction}

Semantic segmentation is a fundamental task in computer vision and has been widely applied to autonomous driving, robotics, remote sensing, and scene understanding~\cite{chai2026like, yan2026language}. Recent advances in large-scale pre-trained foundation models have significantly improved segmentation performance on standard benchmarks such as ADE20K~\cite{zhou2019semantic} and Cityscapes~\cite{Cordts2016Cityscapes}. However, despite their strong performance under normal visual conditions, these models often suffer from severe performance degradation when encountering images captured under adverse weather conditions, such as rain, fog, snow, and other natural visual degradations.

Robust semantic segmentation in adverse weather remains a challenging problem. One major difficulty lies in the lack of accurately paired real-world datasets with both clear and weather-degraded images as well as reliable semantic segmentation annotations. Existing datasets often rely on synthetic weather effects or contain spatial misalignments between degraded and clean images, which limits their ability to faithfully evaluate segmentation robustness in real-world weather scenarios. To address this issue, the WeatherProof dataset provides paired clear and weather-degraded images with semantic segmentation labels, enabling a more controlled and realistic benchmark for weather-robust semantic segmentation.

In this technical report, we present our solution for the CVPR 2026 8th UG2+ Challenge Track 2: Semantic Segmentation in Adverse Weather. Instead of relying on external datasets, our method only uses the WeatherProof dataset provided by the challenge. We build a semi-supervised segmentation pipeline based on UniMatch V2~\cite{yang2025unimatch}, where labeled clear-weather images are used for supervised training and degraded-weather images are exploited as unlabeled data. In addition, we employ test-time augmentation during inference to further improve the robustness and accuracy of the final predictions.
\begin{figure*}[!tp]
    \centering
    \captionsetup{skip=0.5pt}
    \includegraphics[width=\linewidth]{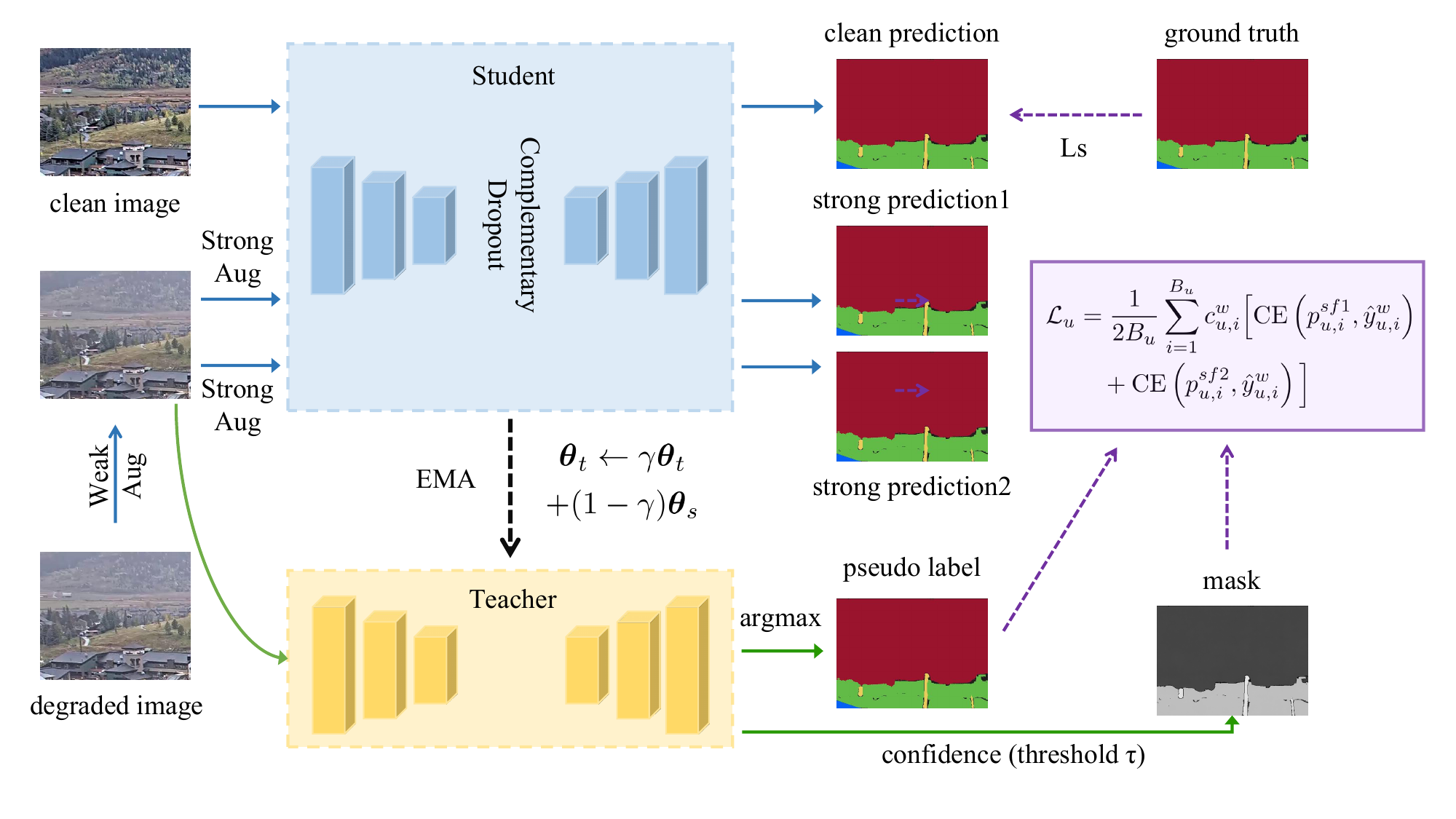}
    \caption{
    Overview of the adopted semi-supervised learning framework. 
    The clean images are fed into the online student network and supervised by the ground-truth masks with a supervised loss. 
    For degraded images, a weakly augmented view is processed by the EMA teacher to generate pseudo labels and confidence masks, while two strongly augmented views are fed into the student network with complementary dropout to produce strong predictions. 
    The unlabeled loss is computed by matching the strong predictions with the high-confidence pseudo labels, and the teacher is updated by the exponential moving average of the student parameters.
    }
    \label{fig:framework}
\end{figure*}

\section{Method}

\subsection{Overview}

As illustrated in Fig.~\ref{fig:framework}, we adopt a semi-supervised teacher-student framework to exploit both clean images and degraded images for robust semantic segmentation. 
The key motivation is that clean and degraded images essentially describe the same scene under different weather or degradation conditions. 
Although their visual appearances can be significantly different, their underlying semantic structures should remain consistent. 
Therefore, we treat clean images with annotations as the supervised branch and degraded images without annotations as the consistency branch, encouraging the model to learn weather-invariant semantic representations.

Let the clean image set be denoted as
\[
\mathcal{D}_{c}
=
\{(x_i^{c}, y_i^{c})\}_{i=1}^{N_c},
\]
where \(x_i^{c}\) is the \(i\)-th clean image and \(y_i^{c}\) is its corresponding ground-truth segmentation mask. 
The degraded image set is denoted as
\[
\mathcal{D}_{d}
=
\{x_i^{d}\}_{i=1}^{N_d},
\]
where \(x_i^{d}\) represents a degraded image under adverse weather or appearance degradation.

The segmentation model consists of an encoder \(g\) and a decoder \(h\). 
Following the recent success of large-scale visual representation learning, we adopt DINOv2 as the backbone encoder. 
Specifically, we maintain an online student network
\[
f_s = h_s \circ g_s,
\]
and an exponential moving average (EMA) teacher network
\[
f_t = h_t \circ g_t.
\]
The student network is optimized by gradient back-propagation, while the teacher network is not directly optimized by the loss. 
Instead, the teacher parameters are updated by the EMA of the student parameters.

\subsection{Supervised Learning on Clean Images}

For clean images, the online student network directly predicts the semantic segmentation map:
\[
p_i^{c}
=
f_s(x_i^{c})
=
h_s(g_s(x_i^{c})),
\]
where
\[
p_i^{c}\in \mathbb{R}^{C\times H\times W}
\]
denotes the predicted probability map, \(C\) is the number of semantic classes, and \(H\) and \(W\) are the spatial height and width.

The clean image prediction is supervised by the ground-truth mask \(y_i^{c}\). 
The supervised loss is formulated as
\[
\mathcal{L}_{c}
=
\frac{1}{B_c|\Omega|}
\sum_{i=1}^{B_c}
\sum_{u\in\Omega}
\mathrm{CE}
\left(
p_i^{c}(u), y_i^{c}(u)
\right),
\]
where \(B_c\) is the batch size of clean images, \(\Omega\) denotes the pixel domain, \(u\) indexes a pixel location, and \(\mathrm{CE}(\cdot,\cdot)\) denotes the pixel-wise cross-entropy loss.

\begin{figure*}[!tp]
    \centering
    \includegraphics[width=0.93\linewidth]{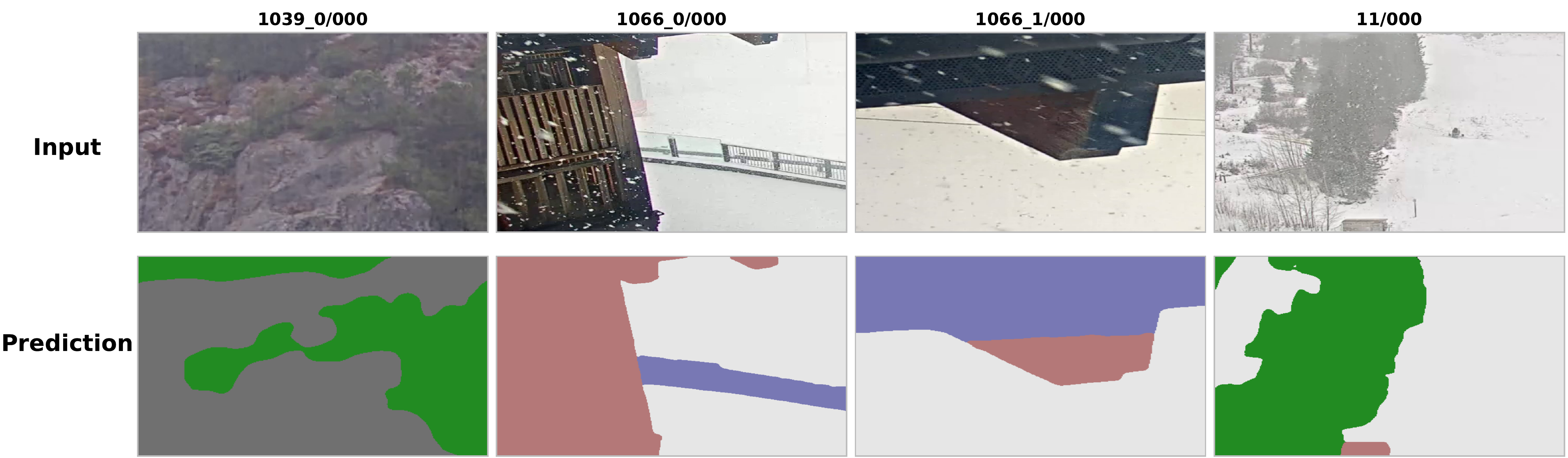}
    \caption{Visualization results of our method.}
    \label{fig:vis_4}
\end{figure*}

\subsection{Weak-to-Strong Consistency on Degraded Images}

For degraded images, we adopt a weak-to-strong consistency learning strategy. 
Given a degraded image \(x_i^{d}\), we first apply weak augmentation to obtain a weakly augmented view:
\[
x_i^{w}
=
A_w(x_i^{d}),
\]
where \(A_w(\cdot)\) denotes weak augmentation. 
In practice, weak augmentation usually contains mild spatial transformations, such as random resizing, cropping, and horizontal flipping. 
Since the weak view preserves most of the original visual content, it is used by the EMA teacher to generate stable pseudo labels.

The weakly augmented degraded image is fed into the teacher network:
\[
p_i^{w}
=
f_t(x_i^{w}),
\]
where
\[
p_i^{w}\in \mathbb{R}^{C\times H\times W}
\]
is the teacher prediction on the weak view.

The pseudo label is obtained by applying the argmax operation along the class dimension:
\[
\hat{y}_i^{w}(u)
=
\arg\max_{c}
p_{i,c}^{w}(u),
\]
where \(p_{i,c}^{w}(u)\) denotes the predicted probability of class \(c\) at pixel \(u\).

To reduce the influence of noisy pseudo labels, we further compute a confidence mask:
\[
c_i^{w}(u)
=
\mathbf{1}
\left(
\max_{c} p_{i,c}^{w}(u)
\geq \tau
\right),
\]
where \(\tau\) is the confidence threshold and \(\mathbf{1}(\cdot)\) is the indicator function. 
Here, \(\hat{y}_i^{w}\) serves as the pseudo semantic target, while \(c_i^{w}\) determines whether each pixel is reliable enough to participate in the unsupervised loss.

\subsection{Strong Augmentation and Complementary Dropout}

To enforce the student network to learn robust representations under stronger perturbations, we generate two strongly augmented views from the weak view:
\[
x_i^{s1}
=
A_s^{1}(x_i^{w}),
\qquad
x_i^{s2}
=
A_s^{2}(x_i^{w}),
\]
where \(A_s^{1}(\cdot)\) and \(A_s^{2}(\cdot)\) are two independently sampled strong augmentations. 
Compared with weak augmentation, strong augmentation introduces stronger appearance and spatial perturbations, such as color jittering, grayscale transformation, Gaussian blur, CutMix-like mixing, and stronger cropping. 
These strong views make the prediction task more challenging and encourage the model to learn degradation-invariant semantic representations.

The two strong views are fed into the shared student encoder:
\[
e_i^{s1}
=
g_s(x_i^{s1}),
\qquad
e_i^{s2}
=
g_s(x_i^{s2}),
\]
where \(e_i^{s1}\) and \(e_i^{s2}\) denote the encoded feature maps.

Following the complementary dropout strategy, we sample a binary channel-wise mask
\[
M\in\{0,1\}^{C_e},
\]
where \(C_e\) is the channel dimension of the encoded feature map. 
The two complementary feature representations are obtained as
\[
\tilde{e}_i^{s1}
=
2M\odot e_i^{s1},
\]
\[
\tilde{e}_i^{s2}
=
2(1-M)\odot e_i^{s2},
\]
where \(\odot\) denotes element-wise multiplication. 
The two masks \(M\) and \(1-M\) select complementary channel subsets, forcing the student to make consistent predictions from different feature subspaces. 
The scaling factor \(2\) is used to preserve the expected feature magnitude after dropout.

The complementary features are then passed through the shared student decoder:
\[
p_i^{sf1}
=
h_s(\tilde{e}_i^{s1}),
\qquad
p_i^{sf2}
=
h_s(\tilde{e}_i^{s2}),
\]
where \(p_i^{sf1}\) and \(p_i^{sf2}\) are the two student predictions. 
The superscript \(sf\) indicates that the predictions are produced from both strong image-level augmentation and feature-level complementary dropout.

\subsection{Unsupervised Loss on Degraded Images}

The teacher-generated pseudo label is used to supervise the two strong predictions of the student network. 
Since strong augmentations may contain spatial transformations, the pseudo label and confidence mask are transformed accordingly to ensure pixel-level alignment with the strong views. 
We denote the aligned pseudo labels and confidence masks for the two strong views as
\[
\hat{y}_i^{w1}, \ c_i^{w1}
\quad \text{and} \quad
\hat{y}_i^{w2}, \ c_i^{w2},
\]
respectively. 
When the strong augmentation only contains appearance perturbations, they are equivalent to
\[
\hat{y}_i^{w1}
=
\hat{y}_i^{w2}
=
\hat{y}_i^{w},
\qquad
c_i^{w1}
=
c_i^{w2}
=
c_i^{w}.
\]

The unsupervised loss on degraded images is formulated as
\[
\begin{aligned}
\mathcal{L}_{d}
&=
\frac{1}{2B_d|\Omega|}
\sum_{i=1}^{B_d}
\sum_{u\in\Omega}
\Big[
c_i^{w1}(u)
\mathrm{CE}
\left(
p_i^{sf1}(u),
\hat{y}_i^{w1}(u)
\right)
\\
&\quad+
c_i^{w2}(u)
\mathrm{CE}
\left(
p_i^{sf2}(u),
\hat{y}_i^{w2}(u)
\right)
\Big],
\end{aligned}
\]
where \(B_d\) is the batch size of degraded images. 
In this loss, the pseudo labels \(\hat{y}_i^{w1}\) and \(\hat{y}_i^{w2}\) provide semantic supervision, while the confidence masks \(c_i^{w1}\) and \(c_i^{w2}\) filter out unreliable pixels.

\subsection{Overall Objective}

The final training objective combines the supervised loss on clean images and the unsupervised consistency loss on degraded images:
\[
\mathcal{L}
=
\mathcal{L}_{c}
+
\lambda
\mathcal{L}_{d},
\]
where \(\lambda\) controls the contribution of the degraded-image consistency loss.

During training, only the student network is updated by gradient back-propagation. 
The teacher network is updated by the exponential moving average of the student parameters:
\[
\boldsymbol{\theta}_{t}
\leftarrow
\gamma
\boldsymbol{\theta}_{t}
+
(1-\gamma)
\boldsymbol{\theta}_{s},
\]
where \(\boldsymbol{\theta}_{t}\) and \(\boldsymbol{\theta}_{s}\) denote the parameters of the teacher and student networks, respectively, and \(\gamma\) is the EMA decay factor.

Through this training strategy, clean images provide reliable ground-truth supervision, while degraded images are exploited through teacher-student consistency regularization. 
The EMA teacher generates stable pseudo labels from weakly augmented degraded images, and the student learns to produce consistent semantic predictions under strong image-level perturbations and complementary feature-level dropout.

\section{Datasets}
WeatherProof~\cite{gella2023weatherproof, zhang2023weatherstream} is a paired semantic segmentation dataset designed for evaluating and improving segmentation robustness under adverse weather conditions. 
Each sample consists of a clear-weather image and its corresponding weather-degraded image, where the two images share nearly the same underlying scene structure and semantic layout. 
This accurate pairing allows the model to learn weather-invariant representations, since the semantic labels should remain consistent while the visual appearance changes due to rain, fog, snow, or other weather effects. 
The dataset contains over 174K images, including 147.8K images for training and 26.2K images for testing. 
The image pairs are selected from GT-RAIN and WeatherStream, and cover diverse geographic locations, urban and natural scenes, camera parameters, and resolutions. 
WeatherProof provides annotations for 10 semantic classes, including background, tree, structure, road, terrain-snow, terrain-grass, terrain-other, stone, building, and sky.

\section{Experiments}

\subsection{Implementation Details}
We conduct experiments on the WeatherProof dataset, which contains 10 semantic classes. 
The original training set is randomly split into training and validation subsets with a ratio of 9:1. 
All input images are randomly cropped to \(518 \times 518\) pixels during training. 
We adopt DPT as the segmentation framework and use DINOv2-Base as the backbone encoder. 
To preserve the general representation capability of the pretrained foundation model, the DINOv2 backbone is frozen during training, and only the task-specific segmentation components are optimized. 
The model is trained for 60 epochs on two NVIDIA V100 GPUs, with a batch size of 16 per GPU, resulting in a total batch size of 32. 
The initial learning rate is set to \(5\times10^{-6}\), and a learning rate multiplier of 40.0 is applied to the trainable segmentation head. 
We use the standard cross-entropy loss for supervised training, where pixels with the ignore label 255 are excluded from loss computation. 
For the semi-supervised branch, the confidence threshold for pseudo-label filtering is set to 0.95, so that only high-confidence teacher predictions are used to supervise the student model.

\subsection{Ablation study}

\begin{table}[t]
\centering
\caption{Ablation study on WeatherProof.}
\label{tab:ablation}
\begin{tabular}{lcc}
\toprule
Setting & mIoU & mDice \\
\midrule
Clean only & 0.69 & 0.69 \\
Clean + Degraded & 0.79 & 0.79 \\
Clean + Degraded + TTA & 0.80 & 0.80 \\
\bottomrule
\end{tabular}
\end{table}

Table~\ref{tab:ablation} reports the ablation results of different training settings on the WeatherProof dataset. 
When only clean images are used for fully supervised training, the model achieves 0.69 mIoU and 0.69 mDice, indicating limited robustness under degraded weather conditions. 
By further incorporating degraded images through the semi-supervised learning strategy, the performance is significantly improved from 0.69 to 0.79 on both mIoU and mDice. 
This demonstrates that the teacher-student consistency learning effectively exploits degraded images and helps the model learn more weather-invariant semantic representations. 
Furthermore, applying test-time augmentation brings an additional improvement, increasing mIoU and mDice from 0.79 to 0.80. 
Although the gain is relatively moderate, it shows that TTA can further enhance prediction stability under degraded conditions. 
Overall, these results verify the effectiveness of using degraded images for semi-supervised training and the complementary benefit of test-time augmentation.

\section{Conclusions}
In this report, we explore semi-supervised semantic segmentation on the WeatherProof dataset. 
By leveraging clean images for supervised learning and degraded images for consistency learning, our method improves robustness under adverse weather conditions and achieves an mIoU of 0.80 in the final evaluation.
{
    \small
    \bibliographystyle{ieeenat_fullname}
    \bibliography{main}
}


\end{document}